\documentclass[10pt,twocolumn,letterpaper]{article}

\usepackage{wacv}

\usepackage{times}
\usepackage{epsfig}
\usepackage{graphicx}
\usepackage{amsmath}
\usepackage{amssymb}
\usepackage[export]{adjustbox}
\usepackage{bm}
\usepackage{bbm}
\usepackage{multirow}
\usepackage{booktabs}

\usepackage[dvipsnames]{xcolor}
\newcommand{\inctabcolsep}[2]{\addtolength{\tabcolsep}{#1} #2 \addtolength{\tabcolsep}{-#1}}
\usepackage[shortlabels]{enumitem}

\renewcommand{\paragraph}[1]{\vspace{.5em}\noindent\textbf{#1}}

\definecolor{mygray}{gray}{0.4}

\def\wacvPaperID{612} 

\wacvfinalcopy 

\ifwacvfinal
\def\assignedStartPage{1} 
\fi

\ifwacvfinal
\usepackage[breaklinks=true,bookmarks=false,colorlinks]{hyperref}
\else
\usepackage[pagebackref=true,breaklinks=true,colorlinks,bookmarks=false]{hyperref}
\fi

\ifwacvfinal
\else
\fi

\begin{document}

\title{Evaluation of Correctness in Unsupervised Many-to-Many Image Translation}

\author{Dina Bashkirova\\
Boston University\\
{\tt\small dbash@bu.edu}
\and
Ben Usman\\
Boston University\\
{\tt\small usmn@bu.edu}
\and
Kate Saenko\\
Boston University and MIT-IBM Watson AI Lab \\
{\tt\small saenko@bu.edu}
}

\maketitle

\begin{abstract} 
  Given an input image from a source domain and a guidance image from a target domain, unsupervised many-to-many image-to-image (UMMI2I) translation methods seek to generate a plausible example from the target domain that preserves domain-invariant information of the input source image and inherits the domain-specific information from the guidance image. 
  For example, when translating female faces to male faces, the generated male face should have the same expression, pose and hair color as the input female image, and the same facial hairstyle and other male-specific attributes as the guidance male image.
  Current state-of-the art UMMI2I methods generate visually pleasing images, but, since for most pairs of real datasets we do not know which attributes are domain-specific and which are domain-invariant, the semantic correctness of existing approaches has not been quantitatively evaluated yet.
  In this paper, we propose a set of benchmarks and metrics for the evaluation of semantic correctness of these methods. 
  We provide an extensive study of existing state-of-the-art UMMI2I translation methods, showing that all methods, to different degrees, fail to infer which attributes are domain-specific and which are domain-invariant from data, and mostly rely on inductive biases hard-coded into their architectures.
\end{abstract}


\section{Introduction}

\begin{figure}[ht]
    \centering
    \includegraphics[width=1.\linewidth]{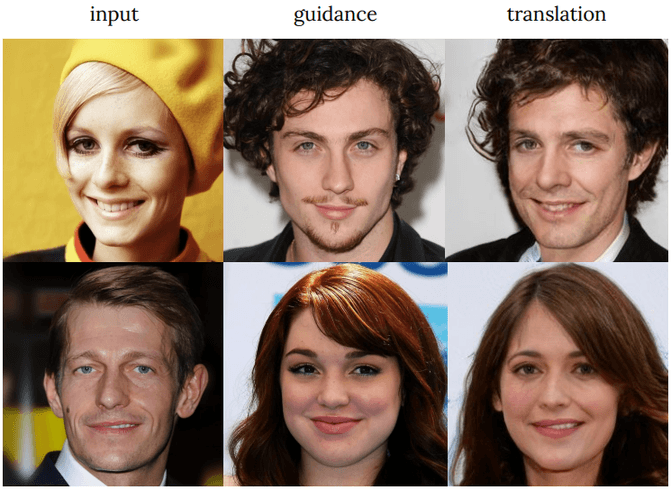}
    \caption{
    Given an input image from source domain and a guidance image from target domain (e.g. male and female photos), unsupervised many-to-many image translation (UMMI2I) methods aim to generate a new image from the target domain with factors varied in both domains (such as pose, facial expression, hair and skin color, etc.) taken from the input image, while the factors that varied only in the target domain (such as facial hairstyle) taken from the guidance image.
    In this paper, we show that current state-of-the-art UMMI2I methods fail to infer which attributes are domain-specific and which are domain-invariant from data. For example, StarGANv2 \cite{choi2020stargan} on the CelebaHQ \cite{CelebAMask-HQ} dataset, falsely considers hair color a domain-specific variation and
    ignores the domain-specific information about facial hair from the guidance image in the top row. 
    We introduce a set of metrics for evaluation of semantic correctness in UMMI2I translation and provide an analysis of current state-of-the-art methods. }
    \label{fig:translation_failure}
\end{figure}

Unsupervised image-to-image translation aims to map an input image from a source domain to a target domain so that the output looks like a valid example of the target domain while preserving the semantics of the input image, given only two sets of source and target images and no supervision. 
First attempts at the unsupervised image-to-image translation \cite{zhu2017unpaired, liu2017unsupervised, kim2017learning, yi2017dualgan, hoffman2018cycada} assumed one-to-one relation between the domains, meaning that for each image in the source domain there exists exactly one corresponding image in the target domain. Unfortunately, this assumption does not hold for most applications, as in most cases there are multiple images in the target domain corresponding to each example in the source domain. This makes the problem ill-posed and results in unreliable translation and unpredictable behaviour on the new unseen examples from the source domain.

\begin{figure*}[ht]
    \centering
    \vspace{-0.5cm}\includegraphics[width=1.98\columnwidth, height=8cm,trim=0.6in 0.3in 0.3in 0.3in,clip]
    {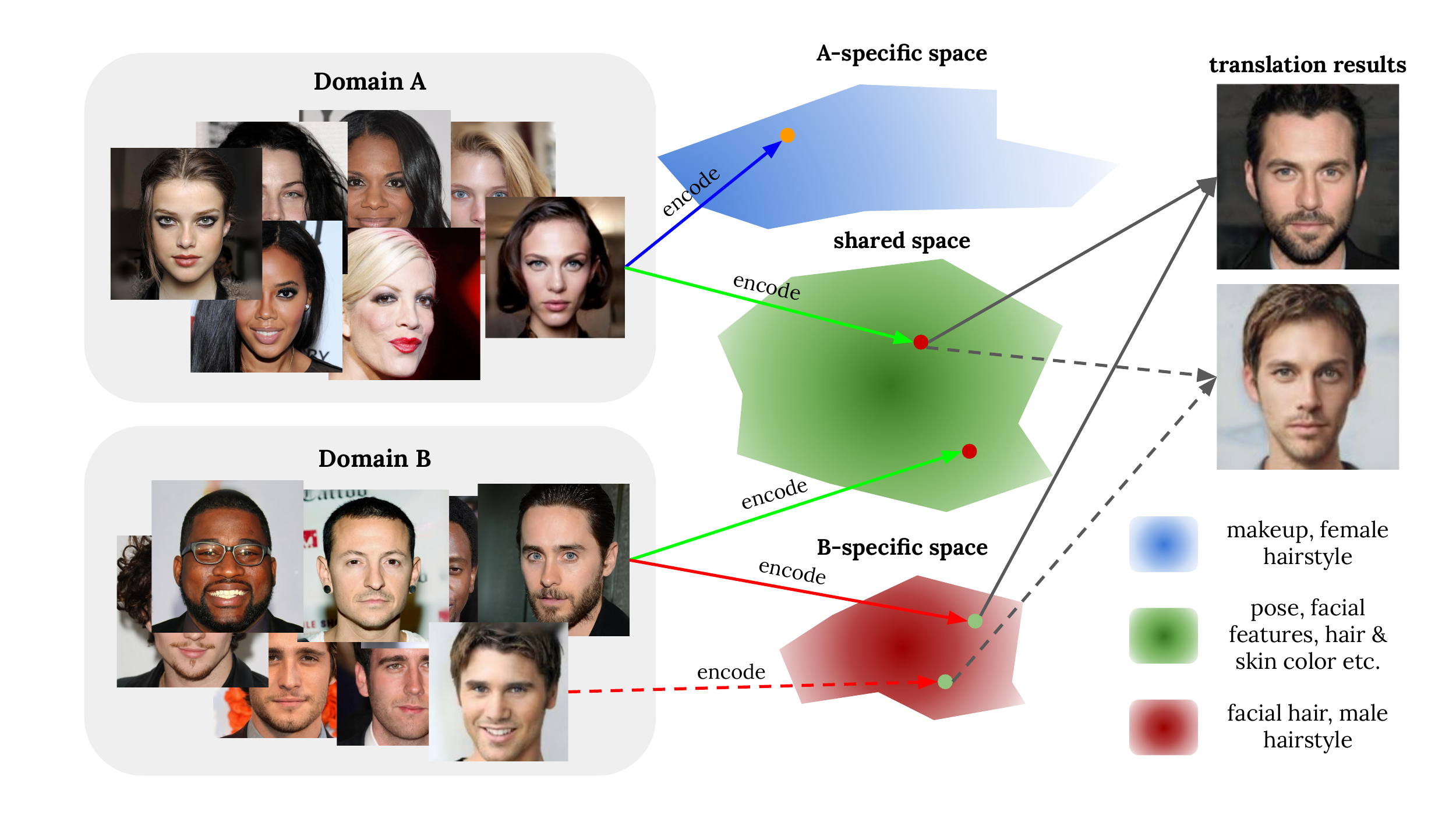}
    \caption{ Overview of the common approach to disentanglement in the UMMI2I translation. Given an input image from the source domain and a guidance example from the target domain, the UMMI2I method produce an image from target domain containing content information from the input image and domain-specific attributes (e.g. facial hair) from the guidance example. 
     A common approach to UMMI2I is to encode images to the domain-specific and domain-invariant embedding spaces. 
  All source examples were taken from CelebaHQ \cite{CelebAMask-HQ} dataset; the translation examples are generated with StarGANv2 \cite{choi2020stargan}
    (\textit{best viewed in color}).}
    \label{fig:fig_1}
\end{figure*}

One natural solution to this problem is to consider many-to-many mappings \cite{huang2018multimodal, ma2018exemplar, almahairi2018augmented, lee2018diverse}, in which case the translation is guided by an example from the target domain that specifies domain-specific factors of the generated target image. For example, when an image of a female face is translated to the male domain, it is not clear whether the translation result should have a beard or not. In the one-to-one setting, the model would either chose the most likely facial hair option or rely on some unrelated property of the input image (e.g. the intensity of the top-left pixel) to make this decision. In the many-to-many setting, however, the information about the desired facial hairstyle is taken from a guidance image from the target domain. The resulting translation problem for a given pair of input and guidance images has a single correct solution, and therefore lets us reason about the correctness of translation models. 
Measuring which architecture choices yield semantically correct UMMI2I translations, \textit{i.e.} provide better disentanglement of domain-specific and domain-invariant factors of variation, is crucial, yet the majority of current state-of-the-art UMMI2I translation methods \cite{huang2018multimodal,almahairi2018augmented,lee2018diverse} are evaluated using only Frechet inception distance (FID) \cite{heusel2017gans}, learned perceptual similarity (LPIPS) \cite{zhang2018unreasonable} and other metrics that measure diversity and realism, and do not take the disentanglement quality into consideration.
Unfortunately, existing disentanglement metrics proposed in the representation learning literature \cite{karaletsos2015bayesian, higgins2016beta, kim2018disentangling, eastwood2018framework, chen2018isolating, kumar2018variational, do2019theory} are not suitable for evaluation of UMMI2I translation either, since they study the quality of disentanglement of latent embeddings, not generated images, and
were designed for single-domain disentanglement, therefore not taking into account
which variations are shared and which are domain-specific. 

In this paper, we propose a new, data-driven approach for evaluation of unsupervised cross-domain disentanglement quality in UMMI2I methods. We designed three evaluation protocols based on the synthetic 3D Shapes  \cite{kim2018disentangling} dataset (originally designed for evaluation of single-domain disentanglement), a more challenging synthetic SynAction \cite{sun2020twostreamvan} pose dataset, and a widely used CelebA \cite{liu2015faceattributes} dataset of faces.
\begin{enumerate}
[wide, labelwidth=!, labelindent=0pt,itemsep=-0.3ex,topsep=0.2ex]
    \item To the best of our knowledge, we are the first to propose a set of metrics for evaluation of the semantic correctness of UMMI2I translation. Our metrics evaluate 
    how well the shared attributes are preserved, how reliably the domain-specific attributes are manipulated, whether the translation result is a 
    valid example of the target domain and
    whether the network collapsed to producing the same most frequent attribute values.
    \item We create three evaluation protocols based on 3D-Shapes, SynAction and CelebA datasets, 
    and measure disentanglement quality of the current state-of-the-art UMMI2I translation methods on them.
    \item 
    We show that for all tested methods there is a clear trade-off between content preservation and manipulation of the domain-specific variations, leading to subpar performance on at least one dataset. More specifically, all methods we tested poorly manipulated attributes associated with adding or changing certain parts of the objects (e.g. facial hair or smile) and AdaIN-based \cite{huang2017arbitrary} methods showed an inductive bias towards treating 
    spatial attributes,
    such as poses and position of objects in the scene, as the domain-invariant factors, and 
    colors and textures as the domain-specific sources of variation, irrespective of which attributes were actually shared between domains and which are domain-specific. 
\end{enumerate}

\section{Related Work}
\subsection{Many-to-Many Image Translation Methods} \vspace{-5px}
\begin{figure*}[ht]
    \centering
    \vspace{-0.8cm}\includegraphics[width=2.\columnwidth,trim=0.1in 1.2in 0.3in 0in]
    {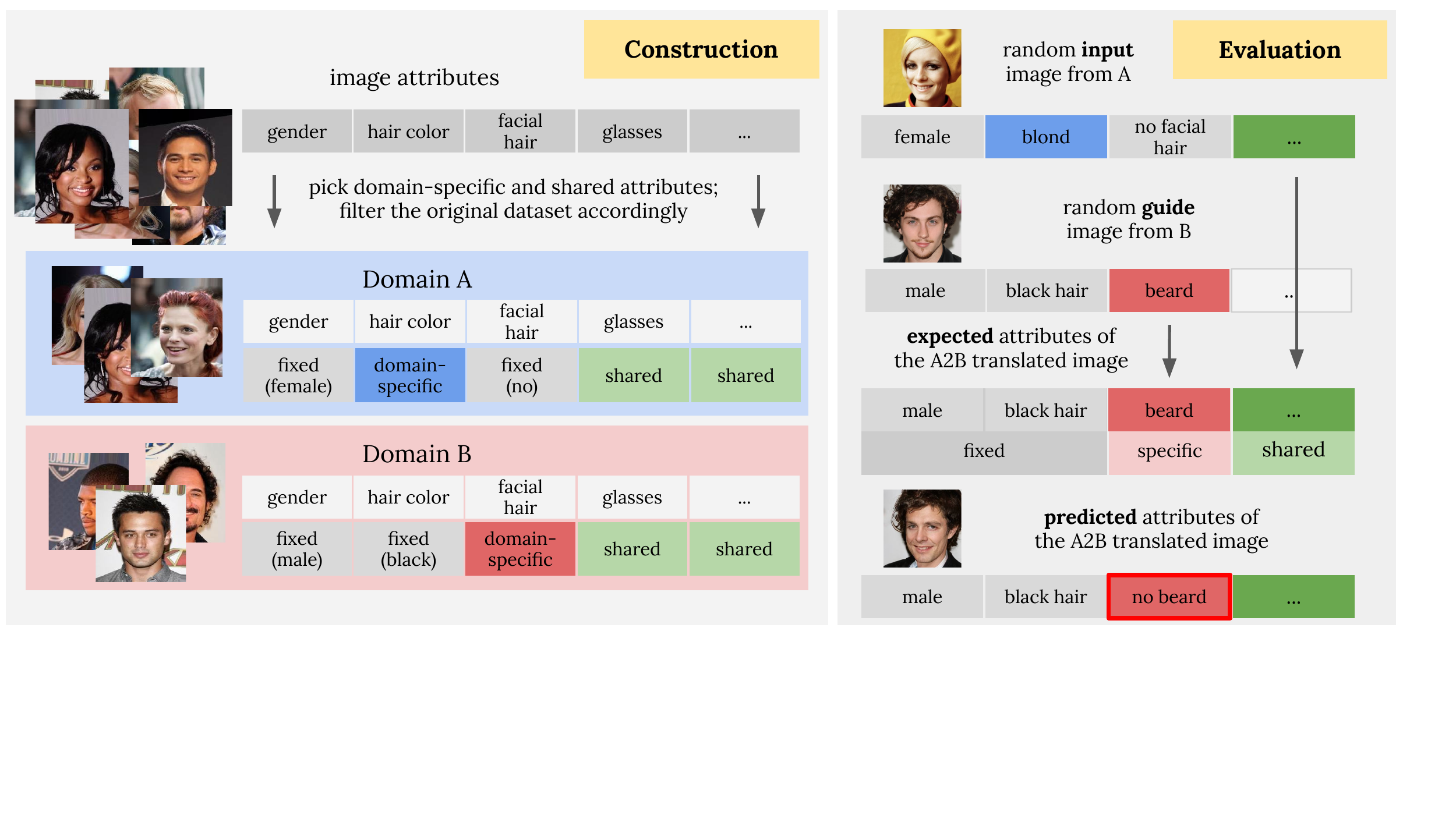}

    \caption{\textbf{Left:} Overview of the disentanglement dataset construction process. Given a dataset of images with some ground truth attribute annotation, we first choose a domain-splitting attribute (e.g. gender); then we choose two non-overlapping sets of the domain-specific attributes (e.g. facial hairstyle for male domain and hair color for female domain); we filter the original datasets so that the domain-specific attributes are varied only in one of the domains (e.g. only kept male photos with black hair). The resulting domains have known domain-specific attributes (hair color vs facial hairstyle) and shared attributes (everything else). 
    \textbf{Right:} To evaluate the disentanglement of the translation example, we compare predicted attribute of the generated images with those of the perfectly disentangled result. The perfectly disentangled translation result has the same content attributes values as the input image, the target domain-specific attributes same as in the guidance image, and the correct values of the fixed attributes (domain-splitting attribute and the opposite domain-specific attributes). }
    \label{fig:fig_2}
\end{figure*}

\paragraph{Style transfer.}  Perhaps one of the first steps towards producing diverse translation results using deep neural networks was the development of neural style transfer methods \cite{gatys2016image, ulyanov2016texture, ghiasi2017exploring, ulyanov2017improved, johnson2016perceptual, zhang2017style, kotovenko2019content}. Conventionally, these methods perform texture transfer by matching the statistics of output features of the input image at a particular layer of a pretrained CNN with those of the guidance image. While style transfer methods show remarkable results in generating artistically stylized images, they are not suitable for general-purpose UMMI2I, because they, effectively, always treat texture and color palette as domain-specific attributes, which is often not sufficient to minimize the domain shift and does not guarantee content preservation. 

\paragraph{Embedding reconstruction.} Another line of work applied reconstruction losses to latent embeddings for disentanglement of domain-specific and shared information.
    For example, Almahairi et al. \cite{almahairi2018augmented} proposed an augmented version of CycleGAN \cite{zhu2017unpaired} with two additional domain-specific encoders. The disentanglement is enforced by the reconstruction loss on latent representations of pairs of source and target examples from the corresponding translation results. 
    The disentanglement strategy introduced in DRIT \cite{lee2018diverse}, DRIT++ \cite{DRIT_plus} and DMIT \cite{yu2019multi}, involves encoding images from both domains to the shared latent space (content space) in addition to the domain-specific encoding. Additionally, those methods share part of the weights of the content encoders and have an additional content discriminator to force the disentanglement of the shared attributes. 

\paragraph{AdaIN-based manipulation.}
    Other UMMI2I translation methods, such as MUNIT \cite{huang2018multimodal}, FUNIT \cite{liu2019few} for few-shot translation, EGSC-IT \cite{ma2018exemplar} and StarGANv2 \cite{choi2020stargan}, used the Adaptive Instance Normalization (AdaIN)  \cite{huang2017arbitrary} to alter the intermediate representation of a generator with ``global'' (shared across all spatial locations) shift and scale parameters extracted from a guidance example. The AdaIN was originally used for style transfer, thus those methods share an assumption that domain-specific factors of variation are ``global'', \textit{e.g.} texture and color variations. In addition to the AdaIN-based manipulation, these methods use separate 
    encoders for the domain-specific information and fully or partially shared encoders for the domain-invariant information and reconstruction losses for the latent representations. 

\paragraph{Domain-specific vs style attributes.} Contrary to prior work \cite{huang2018multimodal}, we prefer the term ``domain-specific attribute'' over ``style attribute'' to mentally distinguish the task from neural style transfer and highlight that in most applications domain-specific factors go beyond the variation in color palette and texture.
We still use ``content'' and ``shared attribute'' as a shorthand for ``domain-invariant factor''.

\subsection{Translation and disentanglement metrics}\label{subsec:related_metrics}\vspace{-5px}
\paragraph{Realism metrics.}
The metrics commonly used for the evaluation of the UMMI2I methods include Inception score (IS) \cite{salimans2016improved} and Frechet Inception score (FID) \cite{heusel2017gans} that measure the similarity between the distributions of the translated images and the target domain, and learned perceptual similarity score (LPIPS) \cite{salimans2016improved} that measures their visual diversity. Neither of these metrics reflects whether domain-specific attributes were taken from the guidance image and whether shared attributes were correctly preserved.


\paragraph{Disentanglement metrics.}
The many-to-many image translation task combines unsupervised disentanglement with pixel-level domain adaptation. Metrics for measuring the quality of unsupervised disentanglement have been extensively explored in the recent literature on representation learning \cite{karaletsos2015bayesian, higgins2016beta, kim2018disentangling, eastwood2018framework, chen2018isolating, kumar2018variational, do2019theory}.  
Perhaps the closest evaluation protocols to ours were used to measure the performance of FactorVAE \cite{kim2018disentangling}, $\beta$-VAE \cite{higgins2016beta} and by Eastwood et al. \cite{eastwood2018framework} that used toy data with known generative factors along with the learned factor regressor to infer the target factors from the latent embedding and comparing these predictions to the expected values. However, estimating whether attributes can be inferred from \textit{intermediate representations} is not sufficient for the evaluation of UMMI2I since it ignores the generative side of this task - ultimately we want realistic images with properly combined attributes, not embeddings.
Additionally, these metrics were developed for single-domain disentanglement of \textit{individual} factors of variation and are not directly suitable for the evaluation of disentanglement of two \textit{groups} of factors from each other (shared factors from domain-specific ones). In this paper, we extend this attribute regression-based approach to unsupervised cross-domain many-to-many translation.


\paragraph{Human evaluation.} 
For the majority of the datasets used for unsupervised image translation, it might not be immediately obvious which sources of variation are shared and which are domain-specific from visual inspection of several images alone. Therefore, it is often unclear what the semantically correct translation should look like from visual inspection of image inputs. This makes it difficult - if not impossible - to instruct human subjects to identify a semantically incorrect translation without clear instructions on which attributes to compare across which domains, resulting in subjects evaluating realism, diversity, and quality of texture transfer, and not the semantic correctness. On the other hand, if we use a controlled dataset with known attribute variations and explicitly instruct the subjects on which attributes are supposed to be transferred from the content or guidance image, it is possible to infer the disentanglement quality from their responses. If we have a reliable regression model for these attributes, then this evaluation pipeline can be fully-automated, which is exactly what we propose in this paper.
In Section \ref{sec:results}, we report results of a user study that indeed matches our automatic evaluation up to the annotator labeling error.

\subsection{Limitations shared with prior work}
Our evaluation protocol shares some limitations with the existing label-based disentanglement evaluation protocols described above, including being limited to datasets with attribute annotations and only focusing on attributes labeled by humans as ``semantic'' introducing some subjectivity. On the other hand, if a specific attribute that is important for a particular application is not properly disentangled, and it is not reflected by our label-based metric because this relevant attribute is not labeled in the data, we can either manually label that attribute or, at the very least, label pairs of images as having correct or wrong attribute transfer, and use it for evaluation. This suggests that, while possibly inevitably subjective, our label-based protocol can be applied in all cases in which human annotators are capable of, at the very least, labeling pairs of images as having same or different values for the most important attributes - which covers the absolute majority of useful practical applications.
\section{Problem statement}

\label{sec:problem_def}
Consider a set of images $A = \{ a_i\}_{i=1}^{N_A}$ from the source domain $\mathcal{A}$, and a set of images $B = \{b_i\}_{i=1}^{N_B}$ from the target domain $\mathcal{B}$. The main assumption about $\mathcal{A}$ and $\mathcal{B}$ is that images in these domains share some semantic structure but differ visually, i.e. daytime vs nighttime photos, CCTV vs front-facing videos, etc. Moreover, we assume that each domain has some domain-specific factors of variations that are either entirely absent or do not vary in the other domain, e.g. we can assume that beards and mustache are entirely absent in the female domain. 
According to the definition provided in MUNIT \cite{huang2018multimodal} and later used in other UMMI2I methods, the task of unsupervised many-to-many (also referred to as multimodal) image translation (UMMI2I) is to learn two mappings $M_{A2B}: \mathcal{A}, \mathcal{B} \rightarrow \mathcal{B} $ and $M_{B2A}: \mathcal{B}, \mathcal{A} \rightarrow \mathcal{A} $ 
that produce images from the target domain with shared factors matching the input image from the source domain and domain-specific factors matching the guidance image from the target domain. For example, assume that females (domain A) in our dataset have variable amount of makeup, and males (domain B) have variable amount of facial hair, and face expression and orientation vary in both. In this case, the correct $M_{A2B}(a, b)$ should be a male face with face expression and orientation as in the input female image $a$ and facial hair as in the male guidance image $b$, and should not be affected by the amount of makeup in $a$. The degree to which the learned mappings satisfy the UMMI2I definition can by measured by computing the fraction of attributes that were correctly taken from respective images. In the next section we propose a set of metrics that factor the overall attribute correctness into several metrics that shed light on specific failure modes of different methods.

\section{Proposed Evaluation Protocol}
\label{sec:protocol}
In this section, we describe proposed evaluation protocol for the UMMI2I problem defined in Section \ref{sec:problem_def}. It comprises two stages: construction of dataset pairs with known dataset-specific and invariant attributes, and estimation of the disentanglement quality on these datasets. 

\paragraph{Dataset construction.}
We assume that the factors of variation of a given dataset can be approximated by a finite set of attributes. 
Unfortunately, the majority of in-the-wild image-to-image translation datasets (horses2zebras, winter2summer) have no attribute-level annotations beyond domain labels, so it is impossible to quantitatively measure the disentanglement quality using these datasets. 
For this reason, we propose to take existing complex single-domain datasets such as CelebA \cite{liu2015faceattributes}, 3D-Shapes \cite{kim2018disentangling} and SynAction \cite{sun2020twostreamvan}, and split each of them into two domains in a way that ensures that domain-specific and domain-invariant attributes are known by design. The overall idea is to choose two sets of attributes that vary in each domain and filter the original single-domain dataset accordingly. Figure \ref{fig:fig_2} gives a schematic overview of the dataset construction and evaluation processes. 
More formally, we consider a single labeled dataset $\mathcal D = \{(\bm{x}^{(i)}, \bm{y}^{(i)}\}_{i}^{N}$ of $N$ images $\bm{x}$ with $M$ labeled attributes $\bm{y} = (y_{1}, \dots, y_{M})$. 

We split attribute indices $\mathcal Z = (1, \dots, M)$ into four non-intersecting groups: a single \textit{domain-splitting attribute} $z_d$ that is constant within each datasets, but differs across datasets, e.g. gender, a set of \textit{domain-invariant attributes} $\mathcal Z_c$ that vary in both datasets, and two sets of \textit{domain-specific attributes} $\mathcal Z_s^A$ and $\mathcal Z_s^B$ that vary only in respective datasets and are fixed in the other dataset:
\begin{equation}
    \mathcal Z = \{z_d\} \sqcup \mathcal Z_c \sqcup \mathcal Z_s^A \sqcup \mathcal Z_s^B.
\end{equation}
By definition, domain-specific attributes must be fixed in one domain and vary in the other, and the domain-splitting attribute should be fixed in both domains, so in order to fully specify the dataset, we also need to pick fixed values $q^A, q^B$ for the domain-splitting attribute, and values $\{t_k^A\}_{k \in \mathcal Z_s^B}$ and $\{t_k^B\}_{k \in \mathcal Z_s^A}$ for domain-specific attributes in the domain in which these attributes are fixed. This way, the final dataset split can be fully specified as follows:
\begin{gather*}
    A = \{\bm{x} \ | (y_{z_d} = q^A) \wedge (y_{k} = t_k^A), k \in \mathcal Z_s^B, (\bm{x}, \bm{y}) \in \mathcal D\}, \\
    B = \{\bm{x} \ | (y_{z_d} = q^B) \wedge (y_{k} = t_k^B), k \in \mathcal Z_s^A, (\bm{x}, \bm{y}) \in \mathcal D\}.
\end{gather*}
As a result, we can use the subsets $A$ and $B$ as a pair of domains to measure the UMMI2I disentanglement, since their domain-specific attributes $\mathcal{Z}_{s}^A$ and $\mathcal{Z}_{s}^B$ and shared attributes $\mathcal{Z}_{c}$ are known by design.
Therefore, an ideal many-to-many image translation network $M_{A2B}$ trained on a pair of datasets $A, B$ and applied to an input image $a \in A$ and a guidance image $b \in B$ with ground truth attribute vectors $\bm{y}^a, \bm{y}^b$ should yield an image $M_{A2B}(a, b) \in B$ with an attribute vector matching an correct attribute vector $\bm{y^*}$ defined as:
\begin{multline}
y^*_k = 
    y^a_k \cdot \mathbbm{1}[k \in \mathcal{Z}_{c}] + 
    y^b_k \cdot \mathbbm{1}[k \in \mathcal{Z}_{s}^B] +  \\ + 
    t^B_k \cdot \mathbbm{1}[k \in \mathcal{Z}_{s}^A] +
    q^B \cdot \mathbbm{1}[k = z_{d}].
\label{eq:perfect_attr}
\end{multline}

The correct attribute vector for translation in the opposite direction ($M_{B2A}(b, a) \in A$) is defined analogously.

\paragraph{Computing disentanglement metrics.} We start by predicting the attribute vector of the generated image using a pre-trained attribute regression network $R(x)$ as
$\bm{\hat y}(a, b) = R(M_{A2B}(a, b))$.
In practice, unless stated otherwise, to take into account the error of attribute prediction by $R$, we compute the perfect disentanglement attributes $y^*$ using attributes of input images $a$ and $b$ predicted by $R$. In all experiments, we use CNN-based predictors.
A simple agreement metric between $\bm{\hat y}$ and $\bm{y^*}$ fails to differentiate several important failure modes, namely: always generating the input image, or the guidance image, or always generating the same most frequent set of attributes and ignoring input images (\textit{i.e.} mode collapse). To differentiate these failure modes, we introduce following metrics:

\begin{enumerate}[wide, labelwidth=!, labelindent=0pt,itemsep=-0.3ex,topsep=0.2ex]
\item The \textit{overall translation quality} measures whether the translation result semantically belongs to the target domain, but not \emph{not} the visual quality of generated images. It equals to the average probably of the translated image to have the corresponding fixed attribute values of the target domain given that the original input image had different attribute values:
\begin{equation}
    Q^{\text{A2B}}_{\text{tr}} = \mathbb{E}_{k} \operatorname{P}(\hat y_k = y^*_k \mid y^a_k \neq y^b_k), \ k \sim \mathcal{Z}^A_{\text{s}}  \cup \{z_d\}
\end{equation}
\item In order to estimate the \textit{quality of preservation of shared attributes}, we measures the average probability of a shared attribute to be preserved after translation to the target domain given that input images had different values of this attribute:
\begin{equation}
    D_c^{\text{A2B}} = \mathbb{E}_{k} \operatorname{P}(\hat y_k = y_{a}^k \mid y_a^k \neq y_b^k), \ k \sim  \mathcal{Z}_{c} \\ 
\end{equation}
\item We measure the \textit{quality of transfer of domain-specific attributes} similarly:
\begin{gather*}
    D_s^{\text{A2B}} = \mathbb{E}_{k} \operatorname{P}(\hat y_k = y_{b}^k \mid y_a^k \neq y_b^k), \ k \sim  \mathcal{Z}^B_{s}
\end{gather*}
\item Additionally, we can estimate the \textit{model bias for certain attributes} by computing how likely the translation changes an attribute given that it is same for both input and guidance images:
\begin{equation}
    B = \mathbb{E}_{k} \operatorname{P}(\hat y_k \neq y^*_k \mid y_a^k = y_b^k), \ k \sim \mathcal{Z}
\end{equation}
\end{enumerate}

Bias metric $\bm{B}$, which is the probability of the attribute to change if it was same for both the input and guidance image, can be high only in two cases: 1) if a mode collapse appeared and the results are biased towards certain attribute values or 2) if the error of the attribute prediction is high on the translated images. 
We consider attribute predictions for high-bias translations that did not collapse to be low-confidence and report them in \textcolor{mygray}{gray} in tables. These results cannot be used to judge about the disentanglement quality. We also report the average content score $D_c = \frac{1}{2}(D_c^{\text{A2B}} + D_c^{\text{B2A}})/2$ and overall disentanglement quality $D = \frac{1}{4}(D_s^{\text{A2B}} + D_s^{\text{B2A}} + D_c^{\text{A2B}} + D_c^{\text{B2A}})$.

\section{Experiments}
\label{sec:experiments}

\paragraph{Disentanglement datasets.}
We modified 3D-Shapes \cite{kim2018disentangling}, SynAction \cite{sun2020twostreamvan} and CelebA \cite{liu2015faceattributes} to quantitatively evaluate UMMI2I correctness as described above. We note that most attributes can be grouped into one of three kinds: 1) color and texture; 2) shape and position; 3) presence and absence of objects. We split our attributes in a way that tests various kinds of attributes as domain-specific and domain-invariant, and recommend future researchers to follow the same principle on new datasets. 
Precise specifications of attribute splits used in each dataset, and values used for fixed attributes in other domains
can be found in supplementary Section~\ref{sec:sup_datasets} and Table~\ref{tab:celebad_description}. At a high level: 1) \textbf{CelebA-D} (Fig.~\ref{fig:sup_fig_celeba_examples}) uses males with black hair and variable amount of facial hair, and smiling young females with variable hair color as two domains - we opted for this specific split instead of the usual male-female split to have distinct domain-specific attributes of different kinds to each domain (color of hair in one, presence of beards in the other) while keeping domains large enough - there are, unfortunately, not many females in CelebA that are not young and conventionally attractive; 2) \textbf{Shapes-3D} (Fig.~\ref{fig:sup_fig_shapes_examples}) contains 3D renders of one of four shapes in one of ten colors in both dataset, with floor and wall color variations in one domain, and size and view angle variation in the other domain; 3) \textbf{SynAction} (Fig. \ref{fig:sup_fig_synaction_examples}) contains 3D renders of humans in different poses with background texture variations in one domain and clothing and identity variations in the other.

\paragraph{Automatic evaluation.}
We present our results on the three disentanglement datasets described above for Augmented CycleGAN \cite{almahairi2018augmented}, DRIT++ \cite{DRIT_plus}, FUNIT \cite{liu2019few}, MUNIT \cite{huang2018multimodal}, and StarGANv2 \cite{choi2020stargan}. Additionally, we noticed that
the original implementation of MUNIT performs poorly when a the domain-specific embedding is taken from the real guidance image and not sampled randomly. We fixed this issue by adding the domain-specific embedding reconstruction loss and GAN loss for the non-random embedding and reported these results under the name ``MUNITX''.
For all methods, we used the official 
implementations and default hyperparameters. 
To provide a sense of scale, we added four ``naive'' baselines: ``Random triplets'' corresponds to a generator that produces a completely random realistic output image from random domain, ``Random target'' - random realistic output image from the correct domain, ``Content idt'' - always producing the input (content) image, and ``Guidance idt'' -  always producing the guidance image.  We report full lists of attributes, description of the attribute prediction models architectures and reference performance (Fig.~\ref{fig:celeba_attr_hist}-\ref{fig:synaction_pos}), per-attribute disentanglement performance, and more qualitative translation examples (Fig.~\ref{fig:sup_fig_1}-\ref{fig:sup_fig_5}) in the supplementary.

\paragraph{User study.} We ran a blind user study on a random subset of 3D-Shapes asking people to label triplets images as having same or different attribute values (Fig.~\ref{fig:sup_user_study}). We used human responses to these questions instead of predicted attribute values to estimate proposed metrics. Most annotated pairs had one generated and one real image, but a small fraction had two real images with known attributes to estimate the annotation error rate. We described our motivation for this rigid human evaluation protocol in favour of generic ``does $x$ look like $a$ combined with $b$?'' in Section \ref{subsec:related_metrics}.


\section{Results}\label{sec:results}

\begin{figure*}[htb]
    \centering
    \vspace{-0.6cm}\includegraphics[width=2.\columnwidth]{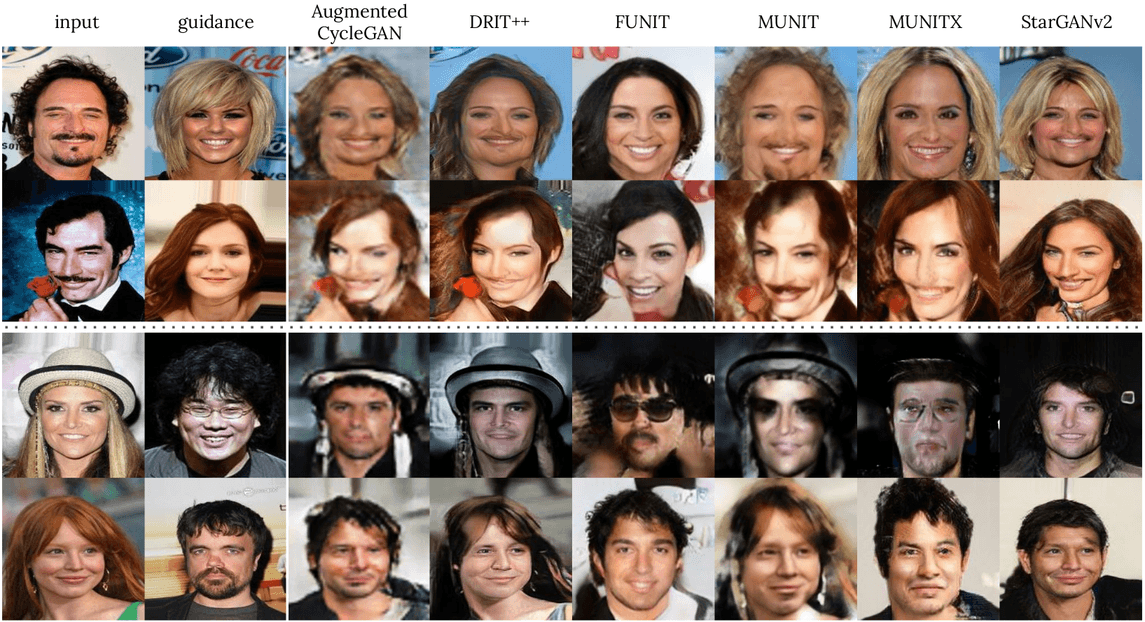}
    \caption{Examples of M$\to$F (top) and F$\to$M (bottom) translations on the proposed \textbf{CelebA-D} subset. A correct translation should have domain-specific attributes of the guidance image (hair color in the top two lines; facial hair, smile and age in the bottom two lines), and the rest of attributes (facial features, orientation, etc.) from the input image. More examples can be found in Figures~\ref{fig:sup_fig_1}-\ref{fig:sup_fig_3} in supplementary. \vspace{-7px}}
    \label{fig:fig_4}
\end{figure*}

\addtolength{\tabcolsep}{-2pt}

\begin{table}[t]
    \centering
    \inctabcolsep{-2pt}{
    \begin{tabular}{lcccccc}\toprule
          \textbf{Model} & $\bm{Q_{\text{tr}}}\uparrow$ & $\bm{D}\uparrow$  & $\bm{D_{s}^\text{\tiny A2B}}\uparrow$ &
          $\bm{D_{s}^\text{\tiny B2A}}\uparrow$ & $\bm{D_{c}}\uparrow$ & $\bm{B}\downarrow$ \\ \midrule
            DRIT++ \cite{DRIT_plus} & $87.8$ & $45.2$ & $49.1$ & $25.1$ & $53.3$  & $12.8$  \\
          MUNIT \cite{huang2018multimodal} & $79.7$ & $44.2$ & $43.0$ & $21.5$ & $\bm{56.2}$ & $\bm{11.1}$ \\
          MUNITX & $81.4$ & $47.4$ & $66.3$ & $20.9$ & $51.2$ & $12.3$ \\
          FUNIT \cite{liu2019few} & $92.1$ & $39.1$ & $31.6$ & $\bm{31.4}$ & $46.7$ & $16.0$\\
          {\small AugCycleGAN} \cite{almahairi2018augmented} & \textcolor{mygray}{$90.9$} & \textcolor{mygray}{$42.1$} & \textcolor{mygray}{$38.5$} & \textcolor{mygray}{$41.3$}& \textcolor{mygray}{$44.2$} & $14.8$ \\
         StarGANv2 \cite{choi2020stargan} & $\bm{94.4}$ & $\bm{50.0}$ & $\bm{80.1}$ & $24.7$ & $47.7$ & $12.2$ \\ 
         \midrule
         Rand. triplets & $72.4$ & $37.5$ & $29.5$ & $20.8$ & $49.9$ & $20.5$ \\
         Rand. target & $96.0$ & $41.6$ 
         & $42.3$ & $37.3$ & $43.4$ & $18.4$ \\
         Content idt & $0.0$ & $50.0$ & $0.0$ & $0.0$ & $100.0$ & $0.0$  \\
         Guidance idt &  $96.0$ & $50.0$ & $100.0$ & $100.0$ &  $0.0$ & $0.0$ \\
         \bottomrule
    \end{tabular}
    }
    \caption{UMMI2I disentanglement metrics on \textbf{CelebA-D} subset. 
    The bottom part contains the results on the ``naive'' baselines described in Section~\ref{sec:experiments}. \textcolor{mygray}{Gray} indicates low confidence in the corresponding disentanglement scores. \vspace{-5px}}
    \label{tab:disent_table_celeba}
\end{table}

\addtolength{\tabcolsep}{2pt}

\begin{figure*}[p]
    \centering
    \vspace{-0.9cm}\includegraphics[width=1.9\columnwidth]{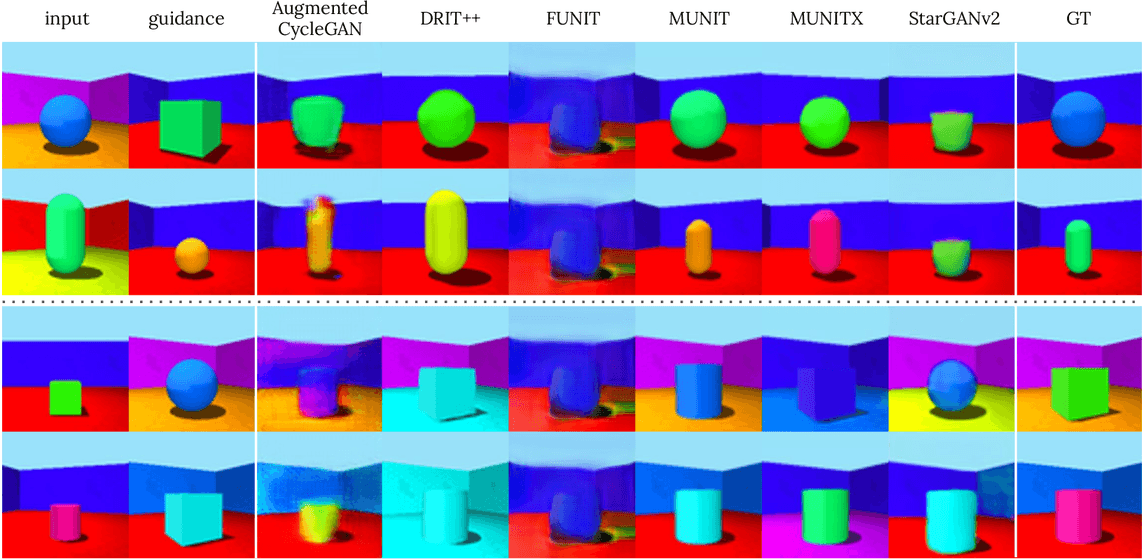}
    \includegraphics[width=1.9\columnwidth,trim=0 0in 0in 2cm,clip]{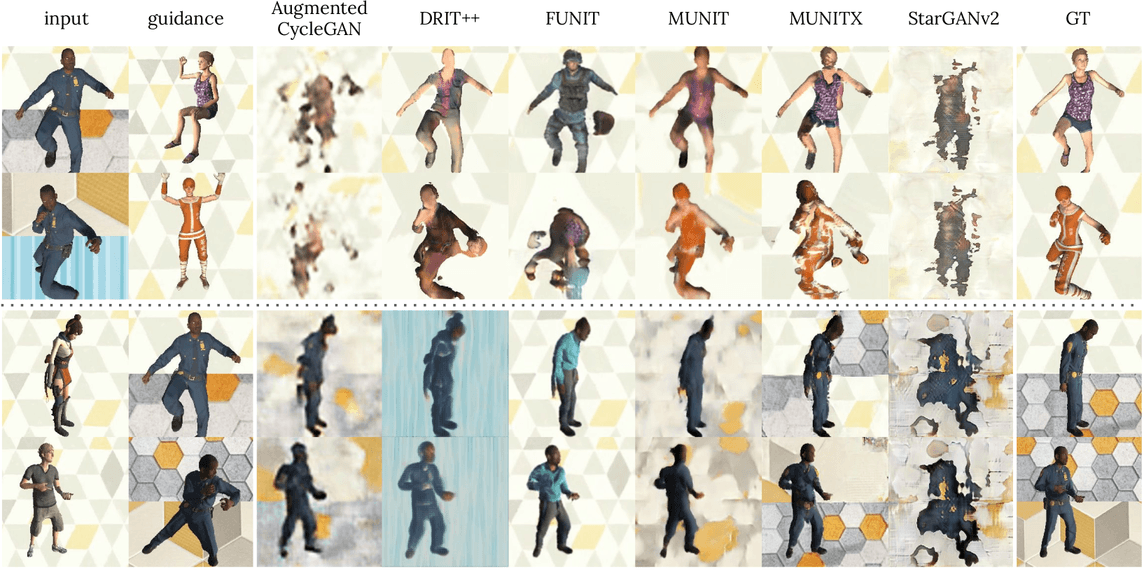}
    \caption{Illustration of UMMI2I translation results on \textbf{3D-Shapes} (top) and \textbf{SynAction} (bottom) subsets. Domain-specific attributes in \textbf{3D-Shapes} are wall and floor color (A) / size and view angle (B), and in \textbf{SynAction} are background texture (A) / clothing and identity (B)}
    \label{fig:fig_5}
\end{figure*}
\begin{table*}[p]
    \centering
    \begin{tabular}{lcccccccccccc}\toprule
    \multirow{2}{*}{\textbf{Model}} & \multicolumn{6}{c}{\textbf{3D Shapes}}  & %
    \multicolumn{6}{c}{\textbf{SynAction}} \\
    \cmidrule(lr){2-7}\cmidrule(lr){8-13}
          & $\bm{Q_{\text{tr}}}\uparrow$ & $\bm{D}\uparrow$  & $\bm{D_{s}^\text{\tiny A2B}}\uparrow$ &
          $\bm{D_{s}^\text{\tiny B2A}}\uparrow$ & $\bm{D_{c}}\uparrow$ & $\bm{B}\downarrow$ &  $\bm{Q_{\text{tr}}}\uparrow$ & $\bm{D}\uparrow$  & $\bm{D_{s}^\text{\tiny A2B}}\uparrow$ &
          $\bm{D_{s}^\text{\tiny B2A}}\uparrow$ &          $\bm{D_{c}}\uparrow$ & $\bm{B}\downarrow$ \\ \midrule
          DRIT++ & $93.5$ & $32.7$ & $10.5$ & $14.9$ & $\bm{52.6}$ & $43.746$ & $89.6$ & $55.7$ & $27.3$ & $12.4$ & $91.6$ &
          $25.7$ \\
          MUNIT & $94.9$ & $\bm{68.1}$ & $\bm{76.3}$ & \bm{$99.7$} & $48.1$ & $\bm{5.12}$ & $86.8$ & $66.9$ & $52.2$ & $19.0$ & $\bm{98.2}$ & $23.8$ \\
          MUNITX & $92.9$ & $33.2$ & $18.6$ & $10.2$ & $52.0$ & $38.2$ 
          & $92.1$ & $\bm{69.4}$ & $\bm{58.3}$ & \bm{$23.7$} & $97.9$ & $20.5$ \\
          FUNIT & $25.0$ & $14.6$ & $18.2$ & $5.39$ & $17.4$ & $66.5$ 
          & $37.4$ & $27.1$ & $6.79$ & $0.5$ & $50.9$ & $\bm{15.2}$ \\
          AugCycleGAN & \textcolor{mygray}{$50.2$} & \textcolor{mygray}{$16.5$} & \textcolor{mygray}{$14.8$} & \textcolor{mygray}{$10.0$} & \textcolor{mygray}{$20.6$} & $63.6$ 
          & \textcolor{mygray}{$99.9$} & \textcolor{mygray}{$37.6$} & \textcolor{mygray}{$14.2$} & \textcolor{mygray}{$36.2$} & \textcolor{mygray}{$50.0$} &
          $33.3$ \\
         StarGANv2 & $\bm{95.9}$ & $29.0$ & $10.3$ & $89.2$ & $8.28$ & $26.0$ 
         & $\bm{96.8}$ & $31.3$ & $17.6$ & $8.69$ & $49.6$ & 
         $33.1$ \\ 
         \midrule
         Rand. triplets & $54.7$ & $11.7$ & $7.77$ & $4.55$ & $17.4$ & $59.4$
          & $49.9$ & $28.0$ & $5.85$ & $6.31$ & $50.0$ &
         $36.5$ \\
         Rand. target & $99.9$ & $14.9$ & $13.5$ & $10.1$ & $18.0$ & $49.5$  
         & $99.9$ & $30.9$ & $11.1$ & $12.7$ & $50.0$ & 
         $18.0$ \\
         Content idt & $0.0$ & $50.0$ & $0.0$ & $0.0$ & $100.0$ & $0.0$ 
         & $0.0$ & $50.0$ & $0.0$ & $0.0$ & $100$ & 
         $0.0$\\
         Guidance idt & $99.8$ & $50.0$ & $100.0$ & $100.0$ & $0.0$ & $0.0$  
         & $99.8$ & $50.0$ & $100$ & $100$ & $0.0$ & $0.0$ 
         \\
         \bottomrule
    \end{tabular}\vspace{10px}
    \caption{The UMMI2I disentanglement evaluation metrics on \textbf{3D-Shapes} and \textbf{SynAction}. 
    The notation is consistent with in Table \ref{tab:disent_table_celeba}. AdaIN-based methods tend to treat color and texture-based attributes as domain-specific even if they are domain-invariant, while non-AdaIN methods preserve domain-invariant attributes better, but fail to modify color and texture even if they are domain-specific attributes.
    }
    \label{tab:disent_table_dshapes_synaction}
\end{table*}


Results in Tables \ref{tab:disent_table_celeba} and \ref{tab:disent_table_dshapes_synaction} as well as the per-attribute results (supplementary Section \ref{sup:tables}) indicate the following:
\begin{enumerate}[wide, labelwidth=!, labelindent=0pt,itemsep=-0.3ex,topsep=0.2ex]
\item AdaIN-based methods (MUNIT, MUNITX, FUNIT and StarGANv2) tend to treat color- and texture- related information as domain-specific factors, even if these factors are actually domain-invariant and should be preserved, and fail to capture spatially localized domain-specific factors. For example, MUNITX and StarGANv2 perform hair color (domain-specific) change correctly with more than a $80\%$ accuracy on average in Celeb-A, and were able to correctly change the wall and floor hue (domain-specific) in Shapes-3D, and MUNITX correctly changed the identity and background (domain-specific) in SynAction. But on Shapes-3D, these methods changed the object hue (domain-invariant) that should have been preserved, failed to correctly change the smile with the accuracy $34\%$ and $22\%$ respectively, or add a correct facial hairstyle (below $28\%$ and $16.6\%$). Also, MUNITX failed to apply the background texture in correct spatial location on SynAction (see Figure \ref{fig:fig_5}).

\item Non-AdaIN-based methods (DRIT++, AugCycleGAN) preserved all domain-invariant attributes better across all experiments (e.g. correctly preserves eyeglasses ($82\%$) and hats ($81\%$) from the content image compared to MUNITX's $54\%$ and $57\%$ respectively), but performed poorly in transferring any domain-specific information including color- and texture- based attributes with the average accuracy of hair color change $\approx 33\%$ and accuracy of correct wall and floor hue change on 3D-Shapes below $16\%$.

\item All methods preserved domain-invariant spatial information well, such as head orientation in CelebA or body pose in SynAction, with the translation pose being closer to that of the input image than the guidance image $\textgreater 91\%$ of cases (except FUNIT and AugCycleGAN - see below). The best models in this aspect are MUNIT and MUNITX ($98\%$ and $76\%$ on CelebA and SynAction respectively).

\item 
On all three datasets FUNIT experiments resulted in moderate to severe mode-collapse, resulting in low correctness and high bias scores. Low visual fidelity of images generated by AugCycleGAN prevented reliable automatic evaluation (metrics reported in gray in all tables).

\item The user study reported in Table~\ref{table:human_eval} in supplementary, resulted in same trends as the automatic evaluation reported in the main paper (Table \ref{tab:disent_table_dshapes_synaction}), confirming that the discrepancy between automatic and human evaluation comes from human annotation error (observed human error rate $<$10\%).

\end{enumerate}

More qualitative results for all methods on all datasets can be found in supplementary Figures~\ref{fig:sup_fig_1}-\ref{fig:sup_fig_5}. Overall, all methods performed well in preserving spatial domain-invariant features, and some performed better then others in transferring global textures and colors, but no single method was able to perform well across all three kinds of domain-specific and global domain-invariant factors of variation.

\section{Conclusion}
We believe that one can only improve upon the aspects of the model that can be quantitatively evaluated, and our work provides a much needed way to measure the semantic correctness of UMMI2I translation. 
In this paper, we formalized the notion of semantic correctness of image translation in terms of shared and domain-specific attributes and proposed three protocols for the evaluation of UMMI2I methods. The results of five state-of-the-art UMMI2I translation methods show that all methods, to different degrees, fail to infer which attributes are domain-specific and which are domain-invariant from data, and mostly rely on inductive biases hard-coded into their architectures.
Consequently, these methods perform well only on a very limited range of applications that match their inductive biases. We conclude that more effort should be made to create UMMI2I methods that enable actual data-driven disentanglement of domain-specific and shared factors of variation. 
\newpage

{\small
\bibliographystyle{ieee_fullname}
\bibliography{egbib}
}
 \clearpage
\section{Supplementary material}
\subsection{More details on the datasets}
\label{sec:sup_datasets}

We did not split the data into train and test subsets to train disentanglement methods in order to exclude the effect of possible differences in the train/test distributions on the disentanglement results. It does not lead to overfitting because the task is unsupervised and we never observe any ground truth during training.

\paragraph{3D-Shapes.}
We modified the 3D Shapes \cite{kim2018disentangling} dataset commonly used to evaluate disentanglement in representation learning. The original dataset contains the following attributes: object shape, object hue, object size, wall hue, floor hue, and orientation. We modified the 3D-Shapes dataset according to the protocol in Section \ref{sec:protocol} as follows:\\ 
\textit{Content attributes:} shape and hue of the foreground object.
\textit{Domain A-specific attributes:} floor hue (fixed to red in domain B), wall hue (fixed to blue in domain B).
\textit{Domain B-specific attributes:} object size (fixed to 5 out of 8 in domain A) and orientation (fixed in domain A to -30).
Due to the very limited number of attributes in this dataset, we omitted the domain-splitting attribute and considered only the fixed attributes for the evaluation of domain translation quality. The resulting domains A and B contain $4000$ and $4800$ images respectively.

\paragraph{SynAction.} The SynAction \cite{sun2020twostreamvan} dataset is a synthetic dataset containing videos of 10 different actors (identities) performing the same set of actions on 5 various backgrounds. We extended the dataset by introducing 5 more backgrounds by cropping and stitching the available backgrounds to make the dataset more balanced. For this dataset, the available attributes are: identity, pose and background. 
We created the disentanglement dataset by assigning the pose as the \textit{shared} attribute, the background as the domain \textit{A-specific attribute} and the identity as the domain \textit{B-specific attribute}. To compare the pose attribute, we count the translation pose attribute as $1$ if its pose is closer to that of the input image and $0$ otherwise. The resulting SynAction dataset contains $6720$ images in the domain A and $7560$ images in the domain B.

\paragraph{CelebA-D.}
\begin{table}
    \centering
    \begin{tabular}{c|cc}
          Attributes & Male & Female \\\hline
          Hair color & fixed (black) & varied
          \\
          Age & varied & fixed (young) \\
          Smile & varied & fixed (yes) \\
          Facial hair & varied & fixed (no) \\
          Makeup & varied & fixed (yes) \\
          Facial attributes* & content & content \\
         \hline
    \end{tabular}\vspace{1em}
    \caption{Short description of the domain attribute splitting used assemble the CelebA-D dataset. *The full list of content attributes can be found in Section \ref{sec:sup_datasets} of the supplementary material.}
    \label{tab:celebad_description}
\end{table}
To perform the evaluation on a more challenging and commonly used image translation dataset, we modified the CelebA \cite{liu2015faceattributes} dataset containing centered photos of celebrities annotated with 40 attributes, such as hair color, gender, age etc. 
First, we chose the domain-splitting attribute to be "Male", i.e. the dataset was split into Male and Female subsets. We chose hair color as the varied attribute for Female domain, and presence or absence of facial hair, smile and age for Male domain; the hair color for Male domain was fixed to black. We considered the attributes associated with facial features, lighting and pose as the shared attributes (see Table~\ref{tab:celebad_description} for a short description and Section \ref{sec:sup_datasets} of the supplementary material for a detailed list of shared attributes). Our modified CelebA-D subset contains $24661$ and $29627$ images in domains A and B respectively.

To remove inconsistency in the original labels of CelebA, we removed the examples for which the hair color is not annotated. Such choice of style attributes is dictated by the aim to leave as many examples from the original dataset as possible, i.e. to filter out the smallest number of examples when the opposite domain style attributes are being fixed, while using the most visible attributes as style attributes. 

The list of content attributes:
``5\_o\_Clock\_Shadow'', ``Arched\_Eyebrows'', ``Bags\_Under\_Eyes'', ``Big\_Lips'', ``Big\_Nose'', ``Blurry'', ``Bushy\_Eyebrows'', ``Chubby'', ``Double\_Chin'', ``Eyeglasses'', ``High\_Cheekbones'', ``Narrow\_Eyes'', ``Oval\_Face'', ``Pale\_Skin'', ``Pointy\_Nose'', ``Straight\_Hair'', ``Wavy\_Hair'', ``Wearing\_Hat''.

\subsection{Attribute prediction networks}
\label{fig:attr_preditors}
If not stated otherwise, the attribute prediction networks are implemented in Tensorflow \cite{tensorflow2015-whitepaper}.

\paragraph{3D-Shapes}
For wall hue, floor hue, object hue, size and shape classification, we used the CNNs with the following architecture: 2D convolution with 16 $3\times 3$ filters followed by a ReLU activation function and max pooling layer with pooling stride $2 \times 2$, another convolution layer with 32 $3\times 3$ with ReLU activation and  $2 \times 2$ max pooling; dropout layer with the drop probability $0.2$, flattening layer, a dense layer with $128$ units and a final dense prediction layer with the number of units equal to the number of classes in the task. The networks are trained with Adam optimizer \cite{kingma2014adam} using a sparse categorical cross-entropy loss for until convergence. All classifiers achieve nearly $100\%$ test accuracy for all tasks.
For the orientation regression task, we use the following architecture: 2D convolution with 32 $3\times 3$ filters followed by a ReLU activation function and max pooling layer with pooling stride $2 \times 2$, another convolution layer with 16 $3\times 3$ with ReLU activation and  $2 \times 2$ max pooling; dropout layer with the drop probability $0.2$, flattening layer, and the dense layer with a single unit for the final prediction. We use the mean squared error loss and Adam optimizer to train the network. The resulting accuracy on the orientation task is $>98\%$ on test set.

\paragraph{SynAction}
To predict the identity and background, we use the following classification network architecture: three convolution layers with 16, 32 and 64 filters $3\times 3$ filters respectively all followed by a ReLU activation function and max pooling layer with pooling stride $2 \times 2$, dropout layer with the drop probability $0.2$, flattening layer, a dense layer with $128$ units and a final dense prediction layer with the number of units equal to the number of classes in the task. The networks are trained with Adam optimizer \cite{kingma2014adam} using a sparse categorical cross-entropy loss for until convergence. The classifiers achieve $>98\%$ test accuracy for both tasks. For pose estimation, we use the pretrained Personlab \cite{papandreou2018personlab} model from Tensorflow Lite (see pose estimation visualization on SynAction in Figure~\ref{fig:celeba_pos}).

\begin{figure*}
    \centering
    \includegraphics[width=1.\linewidth]{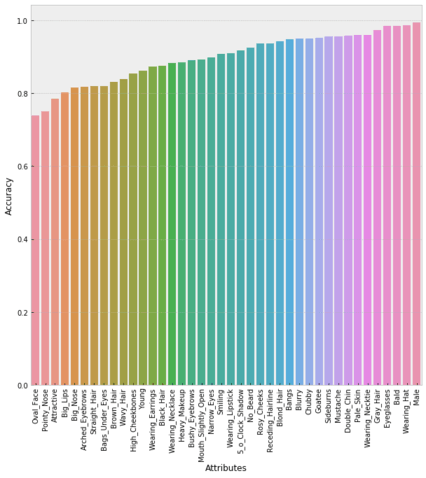}
    \caption{Per-attribute accuracy histogram achieve by our attribute prediction model on \textbf{CelebA} validation split.}
    \label{fig:celeba_attr_hist}
\end{figure*}

\paragraph{CelebA}
For attribute on the CelebA dataset, we used the MobileNetv2 \cite{sandler2018mobilenetv2} feature extractor followed by two dense layers with 1024 and 512 units respectively and ELU non-linearity \cite{clevert2015fast}, and the last dense layer with 40 units and the sigmoid non-linearity. The average attribute classification accuracy of this network is $92\%$, see the detailed information on per-attribute accuracy on Figure \ref{fig:celeba_attr_hist}. Additionally, we measured how well the translation preserves the pose with the  HopeNet~\cite{Ruiz_2018_CVPR_Workshops} model pretrained on the 300W LP dataset~\cite{zhu2016face} and reported the results in Table~\ref{tab:poses_celeba}.

\subsection{Additional Tables}
\label{sup:tables}
Please see Tables~\ref{tab:celeba_content_attrs} and~\ref{tab:celeba_specific_attrs} for attribute-wise disentanglement quality and Table~\ref{tab:poses_celeba} for the pose preservation results on the CelebA-D subset, and Tables~\ref{tab:dshapes_attributes} and~\ref{tab:synaction_attributes} for the attribute-wise disentanglement quality on the 3D-Shapes and SynAction datasets respectively. 

\subsection{Examples of images from generated datasets}
\begin{figure*}[h]
    \centering
    \includegraphics[width=2.\columnwidth]{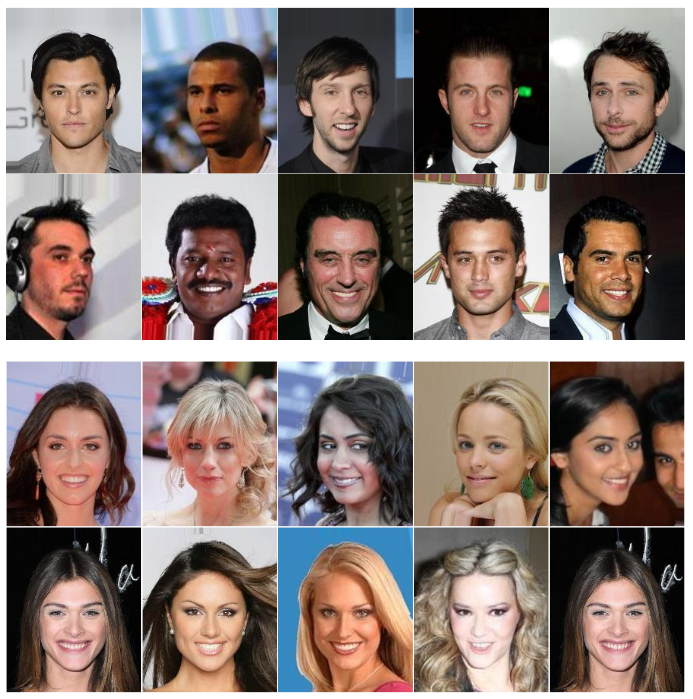}
    \caption{Random examples images from the proposed CelebA split: human faces with variable orientation of 1) males with black hair and variable amount of facial hair, amount of smile and age (top), and 2) smiling young females with variable hair color (bottom). We discuss the motivation behind this specific split in Section 5 of the main paper.}
    \label{fig:sup_fig_celeba_examples}
\end{figure*}

\begin{figure*}[h]
    \centering
    \includegraphics[width=2.\columnwidth]{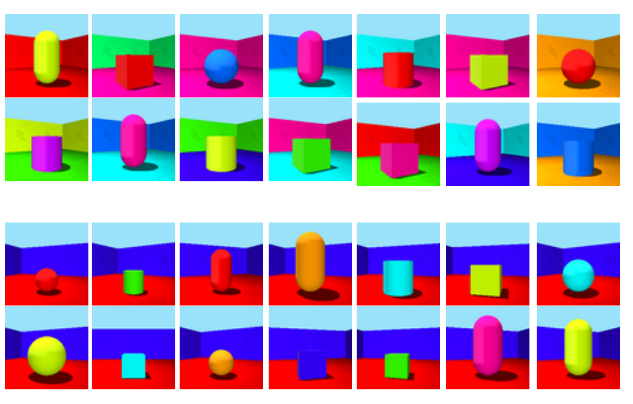}
    \caption{Random examples images from the proposed Shapes-3D split: 3D renders of one of four shapes in one of ten colors in both dataset, with floor and wall color variations in one domain, and size and view angle variation in the other domain}
    \label{fig:sup_fig_shapes_examples}
\end{figure*}

\begin{figure*}[h]
    \centering
    \includegraphics[width=2.\columnwidth]{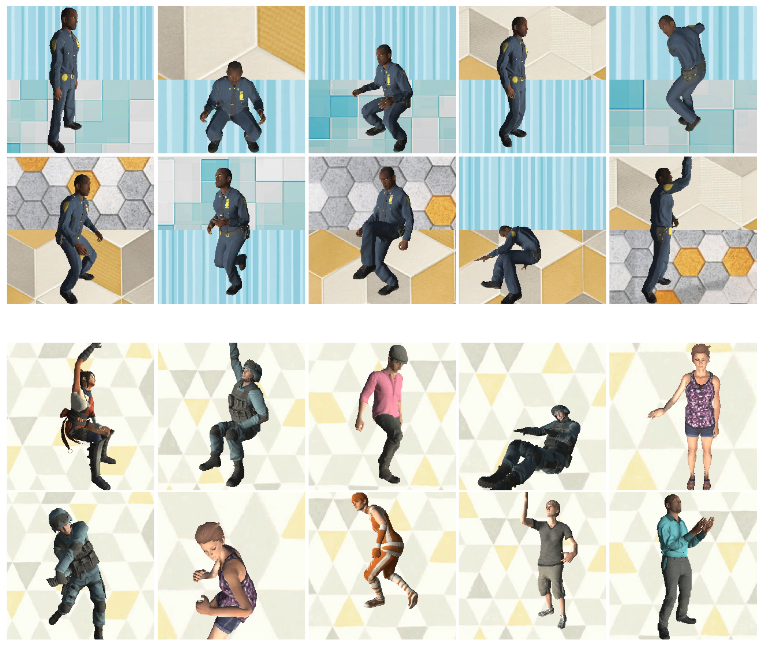}
    \caption{Random examples images from the proposed SynAction split: 3D renders of humans in different poses with background texture variations in one domain (top) and clothing and identity variations in the other (bottom).}
    \label{fig:sup_fig_synaction_examples}
\end{figure*}

For illustrations of generated datasets (CelebA, 3D-Shapes and SynAction) described in details in Section~\ref{sec:sup_datasets} see Figures~\ref{fig:sup_fig_celeba_examples}-\ref{fig:sup_fig_synaction_examples}.

\subsection{More translation examples} 
\begin{figure*}[h]
    \centering
    \vspace{-0.8cm}\includegraphics[width=2.\columnwidth]{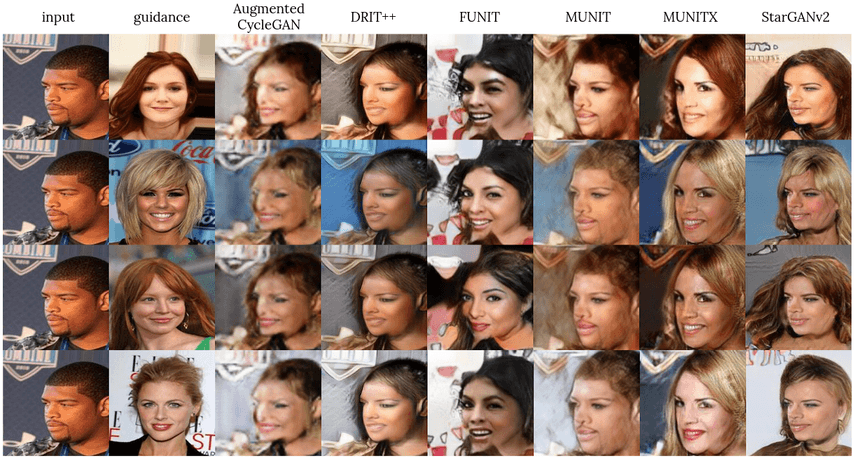}
    \includegraphics[width=2.\columnwidth]{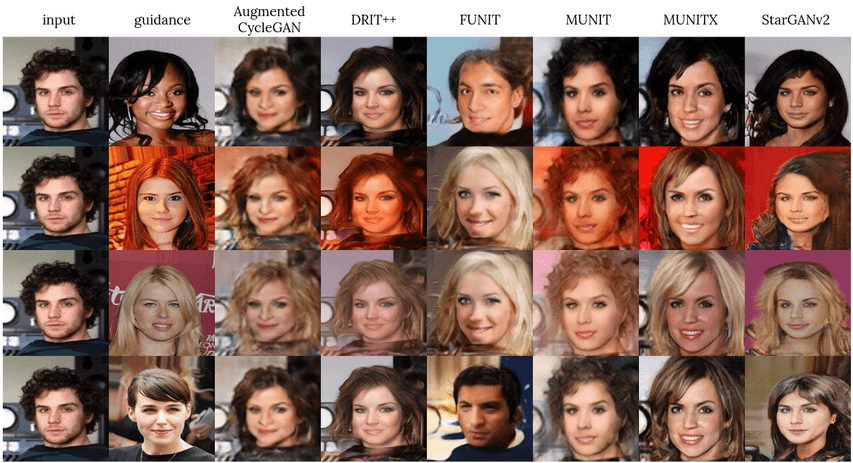}

    \caption{Illustration of many-to-many image translation results on Celeba-D subset. A correct translation should have domain-specific attributes of the guidance image (hair color in the top four lines; facial hair, smile and age in the bottom four lines), and the rest of attributes (facial features, orientation, etc.) from the input image.}
    \label{fig:sup_fig_1}
\end{figure*}

\begin{figure*}[h]
    \centering
    \vspace{-0.8cm}\includegraphics[width=2.\columnwidth]{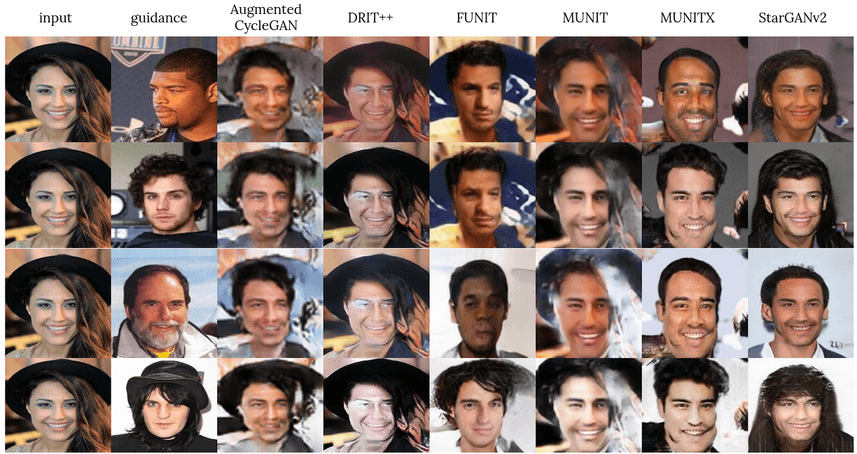}
    \includegraphics[width=2.\columnwidth]{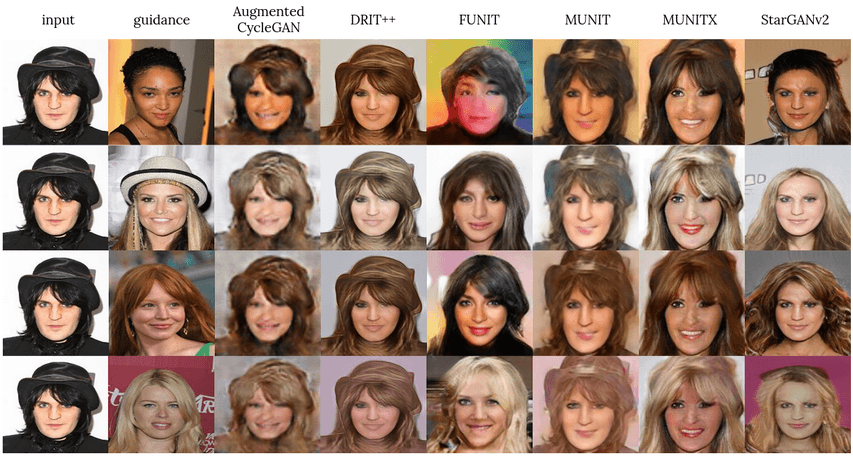}
    \caption{Illustration of many-to-many image translation results on Celeba-D subset. A correct translation should have domain-specific attributes of the guidance image (hair color in the top four lines; facial hair, smile and age in the bottom four lines), and the rest of attributes (facial features, orientation, etc.) from the input image.}
    \label{fig:sup_fig_2}
\end{figure*}

\begin{figure*}[h]
    \centering
    \vspace{-0.8cm}\includegraphics[width=2.\columnwidth]{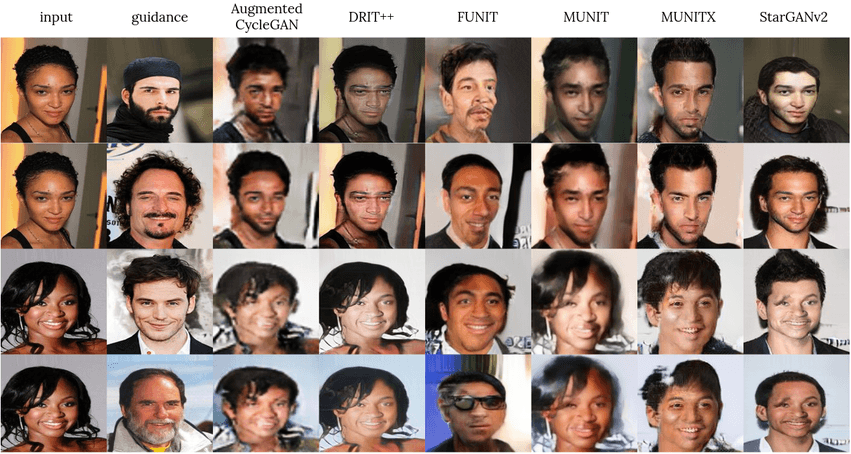}
    \includegraphics[width=2.\columnwidth]{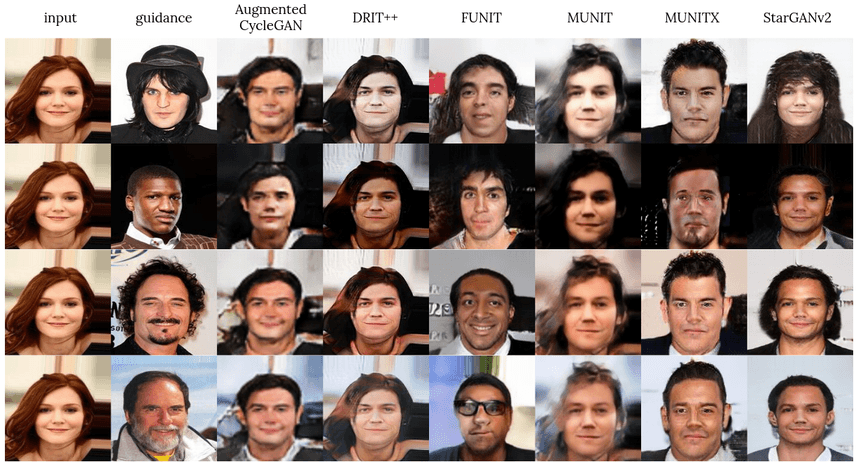}
    \caption{Illustration of many-to-many image translation results on Celeba-D subset. A correct translation should have domain-specific attributes of the guidance image (hair color in the top four lines; facial hair, smile and age in the bottom four lines), and the rest of attributes (facial features, orientation, etc.) from the input image.}
    \label{fig:sup_fig_3}
\end{figure*}

\begin{figure*}[h]
    \centering
    \vspace{-0.8cm}\includegraphics[width=2.\columnwidth]{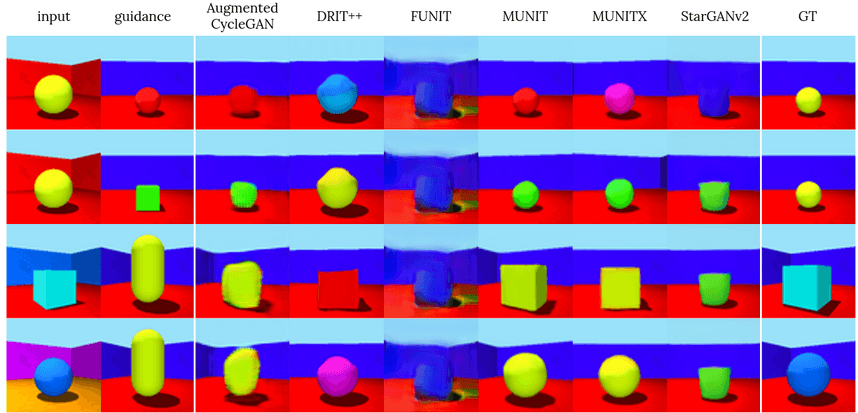}
    \includegraphics[width=2.\columnwidth]{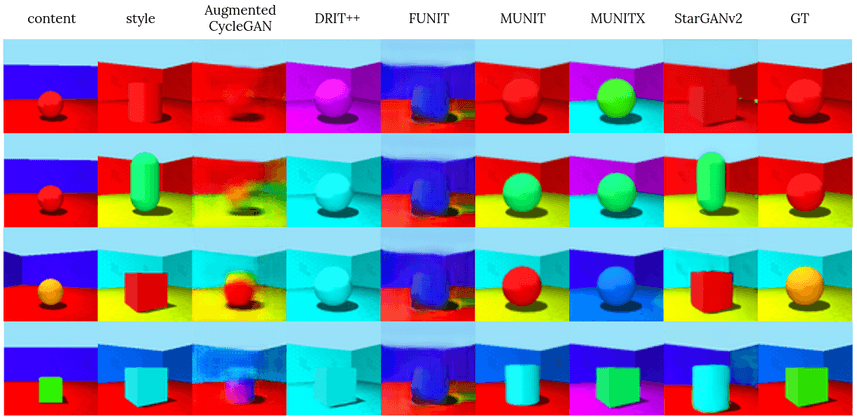}
    \caption{Illustration of many-to-many image translation results on 3D-Shapes subset. A correct translation should have domain-specific attributes of the guidance image (orientation and size in top four lines; wall and floor color in the bottom four lines), and the rest of attributes (shape type and shape color) from the input content image.}
    \label{fig:sup_fig_4}
\end{figure*}

\begin{figure*}[h]
    \centering
    \vspace{-0.8cm}\includegraphics[width=2.\columnwidth]{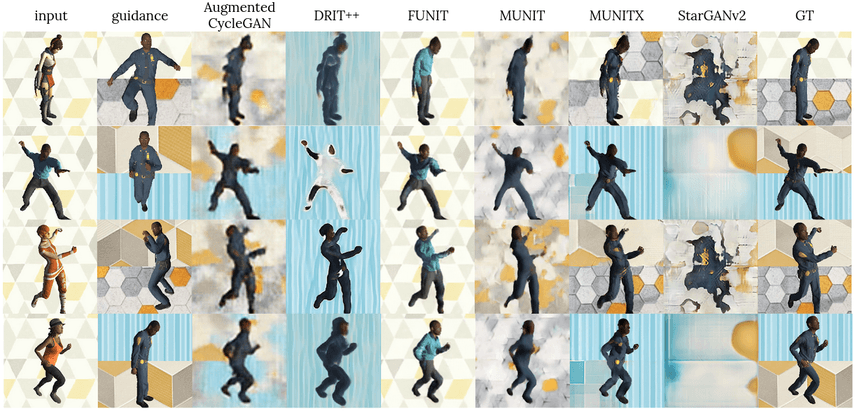}
    \includegraphics[width=2.\columnwidth]{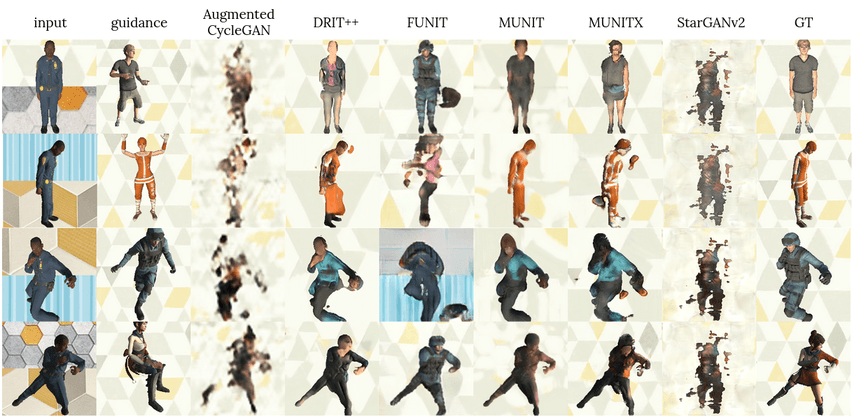}
    \caption{Illustration of many-to-many image translation results on SynAction subset. A correct translation should have domain-specific attributes of the guidance image (background texture in top four lines; clothing and identity in the bottom four lines), and the pose from the input content image.}
    \label{fig:sup_fig_5}
\end{figure*}

For more illustrations of the UMMI2I translation on CelebA, 3D-Shapes and SynAction, please see Figures~\ref{fig:sup_fig_1}-\ref{fig:sup_fig_5}. Our findings are summarized in the Section 6 of the main paper and are backed by metrics introduced and reported in this paper.

\subsection{Pose estimation examples}
\begin{figure*}[h]
    \centering
    \includegraphics[width=1.\linewidth]{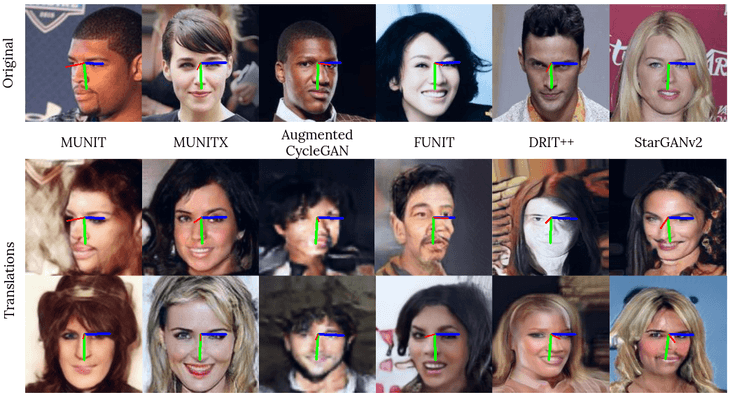}
    \caption{Head pose estimation results on random examples from the original CelebA dataset (\textbf{top}) and random translation results (\textbf{bottom}). Best viewed in color.}
    \label{fig:celeba_pos}
\end{figure*}

\begin{figure*}
    \centering
    \includegraphics[width=1.\linewidth]{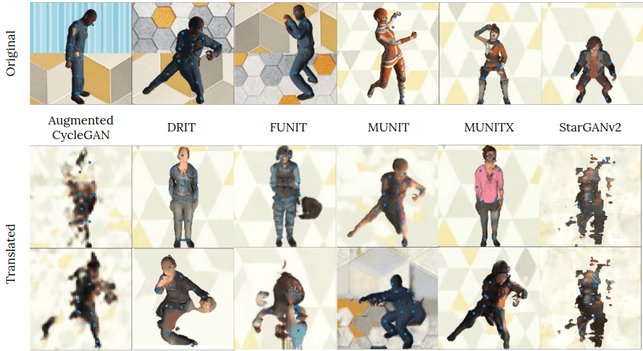}
    \caption{Pose estimation results on random examples from the original SynAction dataset (\textbf{top}) and random translation results (\textbf{bottom}). The pose estimation network succeeded in estimating poses even with severe generation artifacts. Best viewed in color.}
    \label{fig:synaction_pos}
\end{figure*}

Please see the illustration of pose estimation results on Figures~\ref{fig:celeba_pos} and~\ref{fig:synaction_pos}. The pose estimation network succeeded in estimating poses even with severe generation artifacts.

\subsection{User study}
\begin{figure*}[h]
    \centering
    \includegraphics[width=0.5\linewidth]{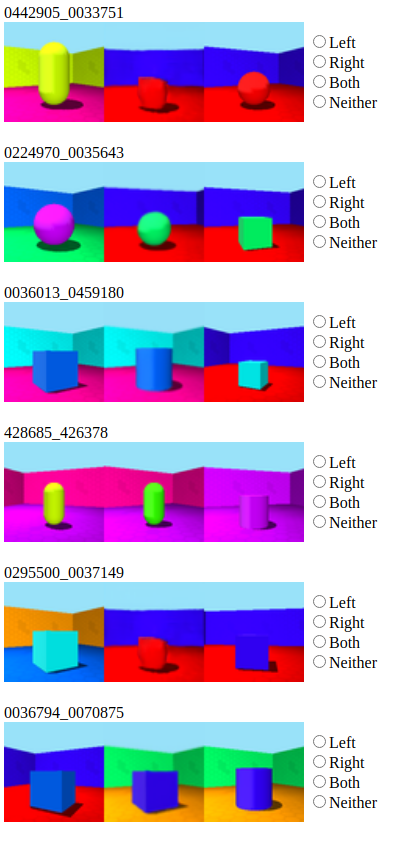}
    \caption{User study screen example. Subjects are explicitly asked to label the \textbf{center} images as having the specific attribute (e.g. shape color or view angle) as coming from either left, right, both images (if both have the same matching attribute value) or neither (if both images have attribute value not matching the center image). These responses were used to compute human evaluation metrics reported in Table~\ref{table:human_eval} and show same trends as automatic evaluation reported in Table~\ref{tab:disent_table_dshapes_synaction}.}
    \label{fig:sup_user_study}
\end{figure*}

Also Table~\ref{table:human_eval} reports results of the human study illustrated and described in Figure~\ref{fig:sup_user_study}. Subjects were explicitly asked to label the images as having the specific attribute matching attribute values other images. These responses were used to compute human evaluation metrics reported in Table~\ref{table:human_eval} and show same trends as automatic evaluation reported in Table~\ref{tab:disent_table_dshapes_synaction}.

\begin{table*}[hp]
    \centering
    \begin{tabular}{c|cccccccccccccccccc}
         \textbf{Method} & 1 & 2 & 3 & 4 & 5 & 6 & 7 & 8 & 9 & 10 & 11 & 12 & 13 & 14 & 15 & 16 & 17 & 18 \\ \hline
         DRIT &   $42$ &  $40$ & $52$ & $\bm{61}$ & $49$ & $54$ & $44$ & $\bm{61}$ & $\bm{59}$ & $82$ & $37$ & $69$ & $38$ & $48$ & $46$ & $\bm{55}$ & $46$ & $\bm{81}$ \\
         
         MUNIT & $50$ & $50$ & $51$ & $58$ & $51$ & $\bm{68}$ & $48$ & $54$ & $53$ & $\bm{89}$ & $\bm{55}$ & $66$& $\bm{45}$ & $46$ & $\bm{64}$ & $51$ & $\bm{51}$ & $65$ \\
         
         MUNITX & $\bm{54}$ & $\bm{53}$ & $\bm{55}$ & $49$ & $55$ & $55$ & $\bm{50}$ & $54$ & $54$ & $54$ & $35$ & $55$ & $44$ & $\bm{49}$ & $53$ & $\bm{55}$ & $43$ & $57$ \\
         
         FUNIT & $52$ & $35$ & $51$ & $51$ & $41$ & $52$ & $41$ & $53$ & $55$ & $52$ & $34$ & $51$ & $38$ & $\bm{49}$ & $39$ & $52$ & $44$ & $54$\\
         
         AugCycleGAN & {\color{gray}$50$} & {\color{gray}$38$} & {\color{gray}$53$} & {\color{gray}$46$} & {\color{gray}$47$} & {\color{gray}$46$} & {\color{gray}$49$} & {\color{gray}$52$} & {\color{gray}$53$} & {\color{gray}$40$} & {\color{gray}$21$} & {\color{gray}$43$} & {\color{gray}$47$} & {\color{gray}$49$} & {\color{gray}$38$} & {\color{gray}$47$} & {\color{gray}$32$} & {\color{gray}$46$} \\
         
         StarGANv2 & $36$ & $40$ & $\bm{54}$ & $50$ & $44$ & $54$ & $44$ & $55$ & $57$ & $78$ & $43$ & $\bm{71}$ & $36$ & $35$ & $49$ & $41$ & $21$ & $51$\\ \bottomrule
    \end{tabular}
    \caption{CelebA content attribute results. The attribute indices correspond to the attributes as follows: 1.``5\_o\_Clock\_Shadow'', 2. ``Arched\_Eyebrows'', 3. ``Bags\_Under\_Eyes'', 4. ``Big\_Lips'', 5. ``Big\_Nose'', 6. ``Blurry'', 7. ``Bushy\_Eyebrows'', 8. ``Chubby'', 9. ``Double\_Chin'', 10. ``Eyeglasses'', 11. ``High\_Cheekbones'', 12. ``Narrow\_Eyes'', 13. ``Oval\_Face'', 14. ``Pale\_Skin'', 15. ``Pointy\_Nose'', 16. ``Straight\_Hair'', 17. ``Wavy\_Hair'', 18. ``Wearing\_Hat''. }
    \label{tab:celeba_content_attrs}
\end{table*}

\vspace{50px}

\begin{table*}[tp]
    \centering
    \begin{tabular}{c|ccc|cccccc}
         \textbf{Method} & Blond & Brown & Black & Young & Smile & Beard & Sideburns & Mustache & Goatee \\ \hline
        DRIT & $31$ & $36$ & $80$ & $20$ & $33$ & $23$ & $\bm{4}$ & $1$ & $2$ \\
        MUNIT & $31$ & $13$ & $86$ & $20$ & $35$ & $6$ & $<1$ & $<1$ & $<1$ \\
        MUNITX & $76$ & $35$ & $87$ & $26$ & $34$ & $16$  & $2$ & $4$ & $\bm{9}$\\
        FUNIT & $28$ & $7$ & $59$ & $22$ & $62$ & $28$ & $1$ & $2$ & $7$\\
        AugCycleGAN & $23$ & $16$ & $76$ & $\bm{32}$ & $\bm{92}$ & $\bm{51}$ & $\bm{4}$ & $\bm{6}$ & $6$\\
        StarGANv2 & $\bm{83}$ & $\bm{68}$ & $\bm{90}$ & $17$ & $22$ & $28$ & $2$ & $2$ & $4$ \\ \bottomrule
    \end{tabular}
    \caption{Per-attribute domain-specific manipulation results on CelebA. Left part: male2female domain-specific attributes (hair color); right part: female2male domain-specific attributes.}
    \label{tab:celeba_specific_attrs}
\end{table*}

\begin{table*}[tp]
    \centering
    \inctabcolsep{-4pt}{
    \begin{tabular}{c|cc|cc|cc}
    
    & \multicolumn{2}{c|}{\textbf{Content}}  & %
    \multicolumn{2}{c|}{\textbf{A-specific}} & \multicolumn{2}{c}{\textbf{B-specific}} \\ \hline
        \textbf{Method} & Obj. hue & Obj. shape & Floor hue & Wall hue & Size & Orientation \\ \hline
         DRIT & $10$ & $95$ & $13$ & $17$ & $14$ & $7$\\
         MUNIT & $<1$ & $\bm{96}$ & $\bm{100}$ & $\bm{100}$ & $\bm{88}$ & $\bm{65}$\\
         MUNITX & $9$ & $95$ & $11$ & $10$ & $29$ & $8$ \\
         FUNIT & $10$ & $25$ & $0$ & $11$ & $29$ & $7$\\
         AugCycleGAN & $\bm{11}$ & $30$ & $10$ & $10$ & $22$ & $8$\\
         StarGANv2 & $5$ & $12$ & $89$ & $89$ & $14$ & $7$\\ \bottomrule
    \end{tabular}
    }
    \caption{Per-attribute results on 3D-Shapes subset. }
    \label{tab:dshapes_attributes}
\end{table*}

\begin{table}[t]
    \centering
    \inctabcolsep{-2pt}{
    \begin{tabular}{c| l l l l l}
          Model & Y $\downarrow$ & P $\downarrow$ & R $\downarrow$ & $D_{p}\downarrow$ & $PM\uparrow$ \\ \hline
          DRIT++ & 3.49 & 4.73 & 1.57 & 3.26 & 0.76 \\
          MUNIT & \textbf{3.17} & \textbf{3.09} & \textbf{1.19} & \textbf{2.51} & \textbf{0.79} \\
          MUNITX & 3.27 & \textbf{3.09} & \textbf{1.19} & \textbf{2.52} & \textbf{0.79}\\
          FUNIT & 5.47 & 6.16 & 1.82 & 4.48 & 0.66 \\
          AugCycleGAN & 16.95 & 8.55 & 3.53 & 9.68 & 0.29 \\
         StarGANv2 & 4.27 & 4.72 & 1.69 & 3.56 & 0.71\\ 
         \hline
         Random Pairs & 20.6 & 9.14 & 3.52 & 11.09 & 0.50 \\
         \hline
    \end{tabular}
    }
    \vspace{10px}
    \caption{Pose preservation metrics for CelebA-D translation with DRIT++\cite{lee2018diverse}, MUNIT\cite{huang2018multimodal}, FUNIT\cite{liu2019few}, Augmented CycleGAN\cite{almahairi2018augmented} and StarGANv2 \cite{choi2020stargan}. The results include mean yaw, pitch and roll distance of the translated image to the content image, overall mean pose distance $D_{p}$ and pose match score $PM$. Distances of random pairs of images are included for comparison. All pose estimation results are estimated by HopeNet \cite{Ruiz_2018_CVPR_Workshops} model. }
    \label{tab:poses_celeba}
\end{table}

\begin{table}[t]
    \centering
    \inctabcolsep{-2pt}{
    \begin{tabular}{c|c|c|c}
    
    & \multicolumn{1}{c|}{\textbf{Content}}  & %
    \multicolumn{1}{c|}{\textbf{A-specific}} & \multicolumn{1}{c}{\textbf{B-specific}} \\ \hline
        \textbf{Method} & Pose & Background & Identity \\ \hline
         DRIT &  $92$ & $13$ & $27$\\
         MUNIT & $\bm{98}$ & $19$ & $52$\\
         MUNITX & $\bm{98}$ & $24$ & $\bm{58}$ \\
         FUNIT & $52$ & $0$ & $7$\\
         AugCycleGAN & $50$ & $\bm{36}$ & $14$\\
         StarGANv2 & $50$ & $9$ & $18$\\
         \hline
    \end{tabular}
    }
    \vspace{10px}
    \caption{Per-attribute results on the SynAction subset. }
    \label{tab:synaction_attributes}
\end{table}

\begin{table}[t]
\centering
\small
\begin{tabular}{cccccc}
\textbf{Method} & $\bm{Q_{\text{tr}}}\uparrow$ & $\bm{D}\uparrow$  & $\bm{D_{s}}\uparrow$ & $\bm{D_{c}}\uparrow$ & $\bm{B}\downarrow$ \\ \hline
DRIT++ & 90.17 & 34.95 & 17.41 & 52.49 & 34.18 \\
MUNIT & 89.86 & 69.16 & 88.44 & 49.88 & 2.12 \\
MUNITX & 89.19 & 41.36 & 28.11 & 54.62 & 39.09 \\
FUNIT & 49.26 & 13.08 & 12.36 & 13.79 & 54.46 \\
AugCycleGAN & 72.81 & 32.33 & 20.97 & 43.69 & 36.73 \\
StarGANv2 & 97.18 & 39.07 & 67.67 & 10.46 & 31.37  \\ 
\hline
\end{tabular} \vspace{10px}
\caption{Human evaluation of disentanglement on 3D-Shapes. It shows same trends as Table~\ref{tab:disent_table_dshapes_synaction} with automatic evaluation results in the main paper.}
\label{table:human_eval}
\end{table}

\clearpage

\end{document}